\documentclass[authoryear, review, 11pt]{elsarticle}
\usepackage{lineno,hyperref}
\usepackage[letterpaper, margin=1in]{geometry}
\modulolinenumbers[5]
\usepackage{booktabs}
\usepackage{threeparttable}
\journal{XX}
\usepackage[table,xcdraw]{xcolor}
\usepackage{multirow}
\usepackage{amsmath}
\usepackage{natbib}
 \bibpunct[, ]{(}{)}{,}{a}{}{,}%
 \def\bibfont{\small}%
\usepackage{algorithmic,algorithm2e,amsmath,xfrac}
\usepackage{setspace}
\usepackage{natbib}
\usepackage{graphicx}
\begin{document}
\title{HeBERT \& HebEMO: a Hebrew BERT Model and a Tool for Polarity Analysis and Emotion Recognition}
\author{Avihay Chriqui, Inbal Yahav}
\date{December 2020}

\begin{abstract}
Sentiment analysis of user-generated content (UGC) can provide valuable information across numerous domains, including marketing, psychology, and public health. Currently, there are very few Hebrew models for natural language processing in general, and for sentiment analysis in particular; indeed, it is not straightforward to develop such models because Hebrew is a Morphologically Rich Language (MRL) with challenging characteristics. Moreover, the only available Hebrew sentiment analysis model, based on a recurrent neural network, was developed for polarity analysis (classifying text as “positive”, “negative”, or neutral) and was not used for detection of finer-grained emotions (e.g., anger, fear, joy). To address these gaps, this paper introduces HeBERT and HebEMO. HeBERT is a Transformer-based model for modern Hebrew text, which relies on a BERT (Bidirectional Encoder Representations for Transformers) architecture. BERT has been shown to outperform alternative architectures in sentiment analysis, and is suggested to be particularly appropriate for MRLs. Analyzing multiple BERT specifications, we find that while model complexity correlates with high performance on language tasks that aim to understand terms in a sentence, a more-parsimonious model better captures the sentiment of an entire sentence. Notably, regardless of the complexity of the BERT specification, our BERT-based language model outperforms all existing Hebrew alternatives on all common language tasks. 
HebEMO is a tool that uses HeBERT to detect polarity and extract emotions from Hebrew UGC. HebEMO is trained on a unique Covid-19-related UGC dataset that we collected and annotated for this study. Data collection and annotation followed an active learning procedure that aimed to maximize predictability. We show that HebEMO yields a high F1-score of 0.96 for polarity classification. Emotion detection reaches F1-scores of 0.78-0.97 for various target emotions, with the exception of \textit{surprise}, which the model failed to capture (F1 = 0.41). These results are better than the best-reported performance, even among English-language models of emotion detection.
\end{abstract}
\maketitle

\section{Introduction}
\textit{Sentiment analysis}, also referred to as opinion mining or subjectivity analysis \citep{liu2012survey}, is probably one of the most common tasks in natural language processing (NLP) \citep{liu2012sentiment,zhang2018deep}. The goal of sentiment analysis is to systematically extract, from written text, what people think or feel toward entities such as products, services, individuals, events, news articles, and topics. 

Sentiment analysis includes multiple types of tasks, one of the most common being \textit{polarity} classification: the binning of overall sentiment into the three categories of positive, neutral, or negative. Another prominent sentiment analysis task is \textit{emotion detection} - a process for extracting finer‐grained emotions such as happiness, anger, and fear from human language. These emotions, in turn, can shed light on individuals’ beliefs, behaviors, or mental states. 
Both polarity classification and emotion detection have proven to yield valuable information in diverse applications. Research in marketing, for example, has shown that emotions that users express in online product reviews affect products' virality and profitability \citep{chitturi2007form,ullah2016valence,adamopoulos2018impact}. In finance, \citet{bellstam2020text} extracted sentiments from financial analysts’ textual descriptions of firm activities, and used those sentiments to measure corporate innovation. In psychology, sentiment analysis has been used to detect distress in psychotherapy patients \citep{shapira2020using}, and to identify specific emotions that might be indicative of suicidal intentions \citep{desmet2013emotion}. Notably, recent studies suggest that the capacity to identify certain emotions (e.g., fear or distress) can contribute towards the understanding of individuals’ behaviors and mental health in the Covid-19 pandemic \citep{ahorsu2020fear, pfefferbaum2020mental}.

The literature offers a considerable number of methods and models for sentiment analysis, with a strong bias towards polarity detection. Models for emotion detection, though less common, are also accessible to the research community in multiple languages. As yet, however, emotion detection models do not support the Hebrew language. In fact, to our knowledge, only one study thus far has developed a Hebrew-language model for sentiment analysis of any kind (specifically, polarity classification; \citet{amram2018representations}). Notably, existing sentiment analysis methods developed for other languages are not easily adjustable to Hebrew, due to unique linguistic and cultural features of this language. 
A key challenge in the development of Hebrew-language sentiment analysis tools relates to the fact that Hebrew is a Morphologically Rich Language (MRL), defined as a language “in which significant information concerning syntactic units and relations is expressed at word-level” \citep{tsarfaty2010statistical}. In Hebrew, as in other MRLs (e.g., Arabic), grammatical relations between words are expressed via the addition of affixes (suffixes, prefixes), instead of the addition of particles. Moreover, the word order in Hebrew sentences is rather flexible. Many words have multiple meanings, which change depending on context. Further, written Hebrew contains vocalization diacritics, known as \textit{Niqqud} (“dots”), which are missing in non-formal scripts; other Hebrew characters represent some, but not all of the vowels. Thus, it is common for words that are pronounced differently to be written in the same way. These unique characteristics of Hebrew pose a challenge in developing appropriate Hebrew NLP models. Architectural choices should be made with care, to ensure that the features of the language are well represented. The current best practice for Hebrew NLP is the use of the multilingual BERT model (mBERT, based on the BERT [Bidirectional Encoder Representations from Transformers] architecture, discussed further below; \citet{devlin2018bert}), which was trained on a small size Hebrew dictionary. When tested on Arabic (the closest language to Hebrew), mBERT was shown to have significantly lower performance than a language-specific BERT model on multiple language tasks \citep{antoun2020arabert}.

This paper achieves two main goals related to the development of Hebrew-language sentiment analysis capabilities. First, we pre-train a language model for modern Hebrew, called \textit{HeBERT}, which can be implemented in diverse NLP tasks, and is expected to be particularly appropriate for sentiment analysis (as compared with alternative model architectures). HeBERT is based on the well-established BERT architecture \citep{devlin2018bert}; the latter was originally trained for the unsupervised fill-in-the-blank task (known as Masked Language Modeling - MLM \citep{fedus2018maskgan}). We train HeBERT on two large-scale Hebrew corpuses – Hebrew Wikipedia and OSCAR (Open Super-large Crawled ALMAnaCH corpus, a huge multilingual corpus based on open web crawl data, \citep{ortiz-suarez-etal-2020-monolingual}. We then evaluate HeBERT’s performance on five key NLP tasks, namely, fill-in-the-blank, out-of-vocabulary (OOV), Name Entity Recognition (NER), Part of Speech (POS), and sentiment (polarity) analysis. We examine several architectural choices for our model and put forward and test hypotheses regarding their relative performance, ultimately selecting the best-performing option. Specifically, we show that while model complexity correlates with high performance on language tasks that aim to understand terms in a sentence, a more-parsimonious model better captures the sentiment of an entire sentence.

Second, we develop a tool to detect sentiments – specifically, polarity and emotions – from user-generated content (UGC). Our sentiment detector, called \textit{HebEMO}, is based on HeBERT and operates on a document level. We apply HebEMO to user-generated comments, from three major news sites in Israel, that were posted in response to Covid-19-related articles during 2020. We chose this dataset on the basis of findings that the Covid-19 pandemic intensified emotions in multiple communities \citep{pedrosa2020emotional}, suggesting that online discourse regarding the pandemic is likely to be highly emotional. Comments were selected for annotation following an innovative semi-supervised iterative labeling approach that aimed to maximize predictability. 

We show that HebEMO achieves a high performance of weighted average F1-score = 0.96 for polarity classification. Emotion detection reaches F1-scores of 0.8-0.97 for the various target emotions, with the exception of \textit{surprise}, which the model failed to capture (F1 = 0.41). These results are better than the best reported performance, even when compared to English-language models for emotion detection \citep{ghanbari2019text, mohammad2018semeval}. 

The remainder of this paper is organized as follows. In the next section, we provide a brief overview of the state of the art in sentiment analysis in general and emotion recognition in particular; we also briefly discuss considerations that must be taken into account when developing pre-trained language models for sentiment analysis. Next, we present HeBERT, our language model, elaborating on how we address some of the unique challenges associated with the Hebrew language. We subsequently describe HebEMO and evaluate its performance on our UGC data. 

\section{Background}
\subsection{Language Specificity in Emotional Expression}
Psychologists and psychoanalysts have long known that, despite the importance of non-verbal behavior, words are the most natural way to externally express an inner emotional world \citep{ortony1987referential}. In line with this premise, theories of emotions stress that emotional experience and its intensity can be inferred from spoken or written language \citep{argaman2010linguistic}. Yet, emotions vary across cultures \citep{rosaldo1984culture}, and, consequently, languages differ in the degree of emotionality they convey and in the ways in which emotions are expressed in words \citep{wierzbicka1994emotion,kovecses2003metaphor}. In particular, as noted by \citet{kovecses2003metaphor}, the verbalization of emotions commonly relies on the use of metaphorical and metonymic expressions, which may differ across languages. Religion is another source of variation in emotional experience and its associated expression \citet{kim2009religion}. One study showed how the moral system of a culture - and specifically, a Middle Eastern culture - can be linked to certain types of emotions, and suggested that differences in culturally dominant emotions can play a decisive role in cultural clashes \citep{fattah2009clash}.

The above discussion implies that emotion detection tools that are implemented in one language might not be easily transferable to other languages, particularly languages that are culturally distant. Accordingly, sentiment analysis tools must be tailored to specific language models in order to provide informative results. The current paper proposes such a tool for the Hebrew language – one that takes into account specific linguistic challenges associated with Hebrew, elaborated in subsequent sections. 

\subsection{Overview of Sentiment Analysis Approaches}
Many studies offer comprehensive overviews of common sentiment analysis methods \citep[e.g.,][]{liu2019survey,hemmatian2019survey,yue2019survey,yadav2020sentiment}. We present here the main points, with an emphasis on models that form the basis of this study. Most of the models described below were developed primarily for polarity analysis; however, as noted in the following subsection, the architectures are applicable to other sentiment analysis tasks such as emotion detection.

Current reviews on sentiment analysis tend to categorize the various approaches according to the granularity level of text that they accommodate \citep{liu2019survey}: \textit{document level}, that is, evaluating whether an entire document expresses a particular type of sentiment (e.g., positive or negative); \textit{sentence level} – assigning a sentiment to each sentence in the document separately; and \textit{aspect level}, that is, assigning sentiment to each “aspect” discussed in the text. The latter requires a pre-processing step to extract aspects from a written text. In this paper we follow a document-level approach, elaborated further below. 

Sentiment classification approaches can further be categorized according to their underlying methodologies. The first, and perhaps the most popular, methodology is the lexicon-based approach. Based on the theory of emotions, this approach uses sentiment terms to score emotions in an input text. \textit{Linguistic inquiry and word count (LIWC)}, for example, is a popular software program that was developed to assess (among other features) emotions in text, using a psychometrically validated internal dictionary \citep{pennebaker2001linguistic}. The main advantage of the lexicon-based approach is that it is \textit{unsupervised}, meaning that it can be applied without any training or labeled data \citep{yue2019survey}. The main limitation of this approach is that is does not account for the context of terms in the lexicon, and thus overlooks complex linguistic features such as sarcasm, ambiguity, and idioms \citep{liu2012sentiment}. Accordingly, its accuracy is fairly low compared with the alternative approaches. 

The second sentiment classification approach is Deep Learning (DL)-based. DL approaches are supervised methods that are based on multiple-layer neural networks. DL-based sentiment classification models differ by their network architecture. Common architectures include the following: (1) \textit{Convolutional Neural Networks (CNNs)}, which transform a structured input layer (e.g., sentences or documents represented as bag-of-words or word-embedding vectors), via convolutional layers, into a sentiment class \citep{kim2014convolutional}; (2) \textit{Recursive or Recurrent Neural Networks (RNNs)}, which handle unstructured sequential data, such as textual sentences, and learn the relations between the sequential elements \citep{dong2014adaptive}; and (3) \textit{Long Short-Term Memory (LSTM)}, a popular variant of RNN, which can catch long-term dependencies between data segments, in one direction (e.g., left to right) or in both (denoted bidirectional LSTM, or \textit{BiLSTM} architecture) \citep{hochreiter1997long}.

In a recent paper, \citet{amram2018representations} raised the question of the relationship between the characteristics of a language and the DL architectural choices of a sentiment classifier. They analyzed this question for the morphologically rich Hebrew language. Specifically, they compared the performance of CNN and BiLSTM architectures on a polarity classification task. They assumed that the latter method would implicitly capture main morphological signatures, and thus outperform the former. Interestingly, and in contrast to findings in English sentiment analysis \citep{yin2017comparative,acheampong2020text}, they found that CNN yielded overall better performance (accuracy = 0.89) than BiLSTM, even when the latter was trained on morphologically segmented inputs. As far as we know, this is the only paper that developed and evaluated a sentiment analysis model for the Hebrew language.

The last sentiment classification method, which we adopt in this paper, is the transfer learning-based approach. Transfer learning is the act of carrying knowledge gained from one problem and applying it to another, similar problem \citep{pan2009survey}. In NLP, transfer learning is implemented via \textit{Transformers} \citep{tay2020efficient}. Similarly to RNN, transformers use a DL approach to process sequential data. The primary advantage of the Transformer is its unique attention mechanism, which eliminates the need to process data in order, and allows for parallelization \citep{vaswani2017attention}. With Transformers, a target language is first algorithmically learned, irrespective of the target language task (e.g., sentiment analysis task). To this end, a language model is trained on a pre-selected unsupervised NLP task (see section \ref{sec:HeBert} for details) . Then the language model is transferred to the target task. This process is called \textit{fine-tuning}. 

Various pre-trained language models have been used in transfer learning for NLP; these include fastText \citep{joulin2016fasttext}, ELMo (Embeddings from Language Models, based on forward and backward LSTMs) \citep{peters2018deep}, GPT (Generative Pre-trained Transformer) \citep{radford2018improving}, and BERT \citep{devlin2018bert}. Of these, BERT is one of the most common Transformer models for NLP. For sentiment analysis tasks, BERT models - and Transformer models in general - are widely used and produce the best results compared with alternatives \citep{zampieri2019semeval, patwa2020semeval}. For the Hebrew language, the only BERT model available is mBERT \citep{devlin2018bert}, which was trained on a small-sized Hebrew dictionary (about 2000 tokens). Notably, for the Arabic language, which is the closest MRL to Hebrew, \citet{antoun2020arabert} showed that a pre-trained Arabic BERT model achieved better performance on polarity analysis than did any other architecture (improvement of 1\% to 6\% in accuracy). The Arabic-specific model also achieved better performance compared with mBERT.

\subsection{Emotion Recognition}
Emotion recognition is a sub-task in sentiment analysis that offers a finer granularity sentiment level compared with polarity analysis. Two definitions of human emotions dominate the NLP literature, with no clear preference between them \citep{kratzwald2018deep}. The first definition, based on a theory developed by \citet{ekman1999basic}, considers emotions as distinct categories, meaning that each emotion differs from the others in important ways rather than simply their intensity. \citet{ekman1999basic} identified six basic emotions, consistent across cultures, that fit facial expressions: anger, disgust, fear, happiness, sadness and surprise. The second definition is based on a theory by \citet{plutchik1980general}, who stressed that emotions can be treated as dimensional constructs, and that there are relations between occurrences and intensities of basic emotions. In particular, \citet{plutchik1980general} defined a “wheel” comprising four polar-pairs of basic emotions: joy--sadness, anger--fear, trust--disgust, and surprise--anticipation. Combinations of dyads or triads of emotions define another set of 56 emotions. For example, \textit{envy} is a combination of \textit{sadness} and \textit{anger}. This wheel serves as the theoretical basis of common automated emotion detection algorithms \citep{medhat2014sentiment}. Notably, for the purpose of emotion detection, the two conceptualizations of emotion are generally compatible with each other, as they agree on the set of emotions defined as “basic” emotions. 

Though common, emotion recognition is not as widespread as polarity analysis, and it is considered more challenging \citep{acheampong2020text}. A key challenge is that, whereas any text can be classified according to its polarity, not all texts contain emotions, and thus it is harder to infer emotions via a lexicon-based approach. This challenge is further compounded by the fact that labeled data are commonly not available. Further, existing datasets are rather imbalanced. Naturally, the lack of data availability is more severe in non-English languages \citep{ahmad2020borrow}.

In general, the emotion detection task is treated as a multi-label classification task, and models for emotion recognition are similar in architecture to polarity detection models. Recent research has shown that, in emotion detection tasks, pre-trained BiLSTM architectures provide advantages over CNN and unidirectional RNN models \citep{acheampong2020text}, and that Transformers are preferable to other DL approaches \citep{chatterjee2019semeval, zhong2019knowledge}. For example, in a recent SemEval competition \citep{chatterjee2019semeval} that included an emotion detection task for three emotions (angry, happy, sad), Transformer-based models were shown to give the best performance (performance ranges: F1-Score = 0.75 - 0.8; precision = 0.78 - 0.85; recall = 0.78 - 0.85). 

\subsection{Training Language Models for Transfer Learning}
\begin{figure}[h]
  \centering
  \includegraphics[width = \columnwidth] {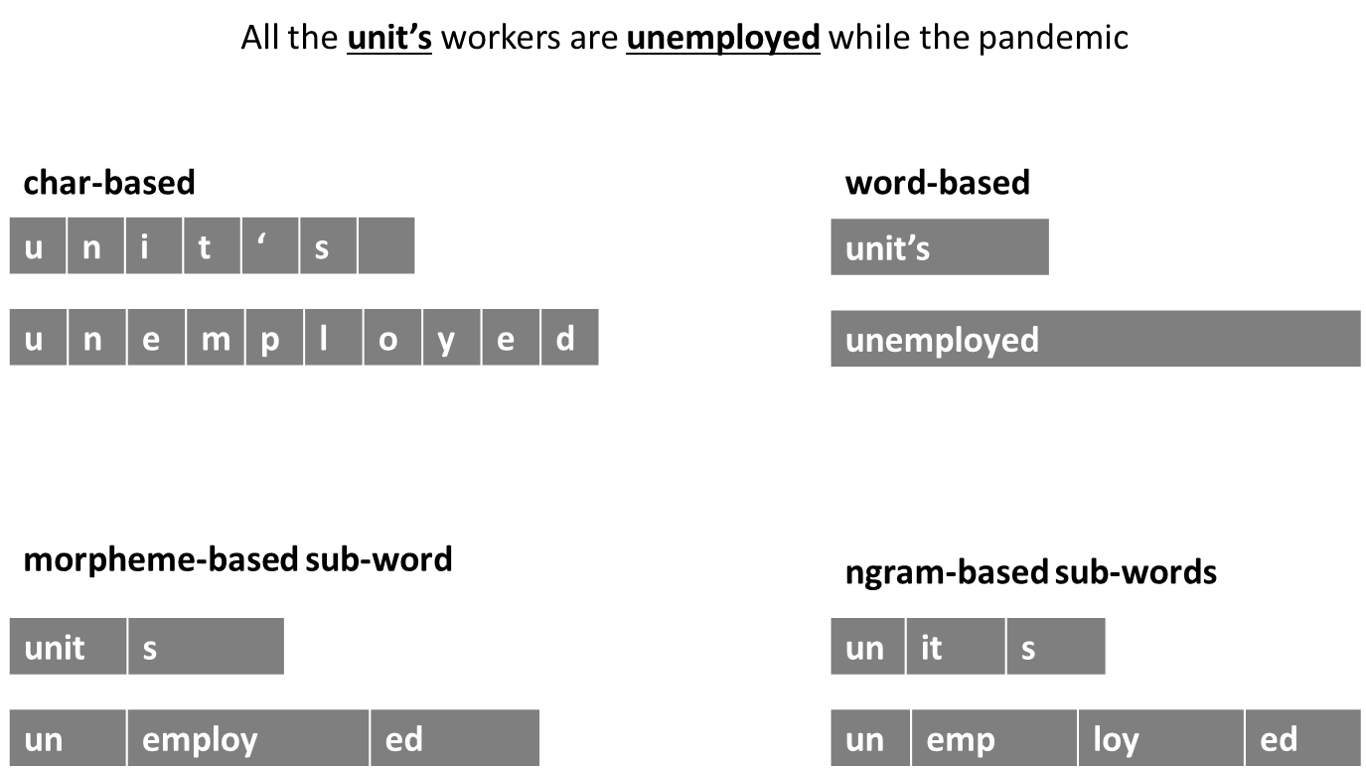}
  \caption{Input representation alternatives}
  \label{fig:compare_tokens}
\end{figure}

As noted above, transfer learning for polarity analysis and/or emotion recognition requires a pre-trained language model. To develop and train a language model, one needs to make the following three basic decisions: 
\begin{enumerate}
  \item Input representation (tokenization): What is the granularity of the tokens that are fed to the model? Common granularity levels include characters, n-gram-based sub-words (using WordPiece algorithm \citep{schuster2012japanese}), morpheme-based sub-words, and full words (see Figure \ref{fig:compare_tokens} for the differences between the approaches).
  \item Architectural choices: What is the exact architecture and specification of the neural network? 
  \item Output: What is the (unsupervised) task that the model is trained on?
\end{enumerate}
Regarding \textit{input representation}, the choice of representation affects the features that the language model is able to capture, and the training complexity. Character-based representation is better for learning word-morphology, especially for low-frequency words and MRLs \citep{belinkov2017neural, vania2018character}, but it comes with longer training time and a deeper architecture, compared with other representations \citep{bojanowski2015alternative}. Word-based representation, in turn, treats each word as a separate token, and thus is considered better for understanding semantics \citep{pota2019multilingual}. With this representation, however, words that differ by prefix or suffix are considered different, necessitating storage of a very large vocabulary. Moreover, out-of-vocabulary (OOV) tokens are not represented. The intermediate option is to use a sub-word representation, which provides some balance between the character- and word-based representations; moreover, it overcomes the OOV problem associated with the word-based representation, and its vocabulary requirements are more manageable \citep{wu2016google}. With sub-words, words can be broken into either n-gram characters, or according to morphemes that have lingual meaning (but also higher computational costs). Previous literature has produced mixed results regarding to the extent to which using a morpheme-based approach can improve upon the n-gram-based approach \citep{bareket2020neural}. Recently, \citet{klein2020getting} showed that sub-word splitting in the multilingual BERT model (mBERT, \citet{devlin2018bert}) is sub-optimal for capturing morphological information.

For the question of \textit{architecture selection}, \citet{devlin2018bert} and \citet{radford2019language} showed that for \underline{similar model size}, BERT outperforms other architectures such as GPT and ELMo on sentiment tasks.

With respect to the \textit{model output}, there are two tasks on which a model can be trained. The first is \textit{predict-the-future}, meaning that the model is trained to predict the last token of a sentence. This task accounts for uni-directional contexts only. The second is the \textit{fill-in-the-blank} task, where the model is trained to fill in a missing token within a sentence. This task takes into account the full (bi-directional) sentence context, and is able to better capture the meanings of tokens, both syntactically and semantically \citep{devlin2018bert}. Recently, \citet{levine2020pmi} offered a method to optimize these tasks, called \textit{Pointwise Mutual Information (PMI) masking}. The authors suggested that instead of filling in a single random token, the model should be trained to fill in a set of tokens that carry mutual information.

\section{HeBERT: Language Model} \label{sec:HeBert}
In this section we develop an unsupervised Hebrew BERT model, which we will later fine-tune for the tasks of polarity analysis and emotion recognition. 
\subsection{Tokenization, Architecture, and Output}
We begin by addressing the three key modeling decisions outlined in the previous section - input representation (tokenization), architecture, and output - in the context of the Hebrew language. 

Recall that, as discussed in the introduction, Hebrew is an MRL with the following important characteristics: (i) grammatical relations in Hebrew are expressed via the addition of affixes; (ii) Hebrew sentences are nearly order-free; (iii) many Hebrew words have multiple meanings, which change depending on context; (iv) Hebrew contains vocalization diacritics that are missing in non-formal scripts, implying that words that are pronounced differently can be written in the same way.

Bearing these features in mind, we first address the \textit{last} two questions, of architectural choice and model output. As discussed in previous sections, BERT has been shown to outperform alternative architectures in sentiment analysis tasks \citep{radford2019language}; moreover, the literature offers evidence that BERT networks effectively capture linguistic information and phrase-level information \citep{jawahar2019does}, a necessary requirement for MRLs \citep{tsarfaty2020spmrl}. Accordingly, we decided to use BERT as our base model, with the default architecture. For the output task, we used BERT’s default \textit{fill-in-the-blank} task. \textit{Fill-in-the-blank} has the advantage of understanding bi-directional context, which corresponds to the order-free property of Hebrew sentences.

With respect to the \textit{input} - the granularity of the tokens - the literature on MRLs, and Hebrew specifically, is inconclusive. \citet{belinkov2017neural} and \citet{vania2018character} showed that character-based representation, which is becoming increasingly popular, is better than word-based representation for learning Hebrew morphology, especially for low-frequency words. For sentiment tasks, however, \citet{amram2018representations} and \citet{tsarfaty2020spmrl} showed that a word-based representation yields better predictions than a char-based representation. With regard to sub-word representations, \citet{klein2020getting} suggested (but did not verify) that, for BERT for Hebrew, morpheme-based sub-words are likely to be preferable to n-gram-based sub-words. A similar argument was made for Arabic, which is the closest MRL language to Hebrew \citep{antoun2020arabert}.

To understand what causes differences in findings between different researchers, consider the following three examples: 
\begin{enumerate}
  \item First, is the word \textit{NA'AL}. \textit{NA'AL} can be translated as either \textit{locked} (e.g., he \textit{locked} the door), \textit{a shoe}, or the past, singular, tense of the verb \textit{wearing} (a shoe). It is also often used as a slang term for \textit{stupid}. The actual semantic meaning of \textit{NA'AL} in a sentence is derived from the context. In that respect, a \textbf{high-level text granularity} (such as a word-based representation) might be the preferable choice for representing Hebrew, as it is better in capturing semantic meanings in context \citep{pota2019multilingual}.
  \item Next, is the word \textit{NA'ALO}, which is an inflection of the word \textit{NA'AL} with the suffix "O". \textit{NA'ALO} can refer to either “his shoe” or “locked it”. In that respect, a \textbf{finer text granularity}, such as char-based, which is better at learning morphology, might be preferred. 
  \item Finally, consider the splitting of the word \textit{NA'ALO}. Here, a meaningful splitting would be \textit{NA'AL-O}. However, such a splitting can be only achieved with \textit{morpheme-based sub-words}, using a tool such as YAP (Yet Another Parser, by \citet{more2019joint}). The alternative, \textbf{n-gram-based sub-words}, will result in additional splitting, which might have lower semantic meaning than morpheme-based sub-words, yet higher robustness to OOV.
\end{enumerate}
Given the above discussion, we hypothesize that sub-word representations (n-gram- or morpheme-based representation), which balance semantic meaning with morphology, will best capture the features of the Hebrew language, and will yield better performance for various language tasks, as compared with character-based and word-based representations. Comparing n-gram-based sub-words with morpheme-based sub-words, we expect the latter to have an advantage on token-level tasks that require a good “understanding” of the language features; yet, a morpheme-based representation might not have such an advantage in document-level downstream tasks. 

To examine our hypothesis, we first train and evaluate multiple small-size BERT models that differ by the granularity of the input. We then choose the best-performing architecture, and re-train the model on a much larger corpus.

\subsection{Comparison Analysis of Tokenization Approaches}
We examine five alternative text representations: char-based; two n-gram-based sub-word representations, which differ in the total vocabulary size (30K tokens vs. 50K tokens); a morpheme-based sub-word representation; and a word-based representation, which considers all words in the corpus, after trimming terms in the lowest $5^{th}$ quantile according to their term frequency (vocabulary size of over 53K tokens).

To compare between the input alternatives, we first train small-sized base-BERTs on a Hebrew Wikipedia dump\footnote{As of September 2013; retrieved from https://u.cs.biu.ac.il/~yogo/hebwiki/. The dataset includes over 63 million words and 3.8 million sentences.}. Our working assumption is that the performance of a small-sized BERT is monotonic with the model’s performance when trained on a larger corpus with the same parameters, yet requires significantly fewer resources. 
We evaluate the models' performances on two common unsupervised language tasks and on three downstream tasks: 
\begin{enumerate}
  \item Unsupervised language tasks:
  \begin{enumerate}
    \item Fill-in-the-blank - the ability to fill in a missing token; tested on a newspaper article \footnote{Retrieved from: https://www.haaretz.co.il} and a fairy-tale dataset\footnote{Retrieved from https://benyehuda.org/ - a volunteer-based free digital library expanding access to Hebrew literature.}. Performance was measured with sequence perplexity ($PP(W)$) - a common measure to examine the ability of a language model to evaluate the correctness of sentences in a sample set. Perplexity of a sequence $W$ with $N$ tokens ($W = \{w_1, w_2, ..., w_n\}$) is calculated as the exponential average log-likelihood of the sequence ($PP(W) = exp\{-\frac{1}{N}\sum_{i}^{N}log_{p_{\theta}}(w_i|w_{<i})\}$, where $log_{p_{\theta}}(w_i|w_{<i})$ is the log-likelihood of the $i^{th}$ token conditioned on the preceding tokens, according to the language model).
    \item Generalizability to OOV - the ability of the language model to generalize beyond the trained corpus (Wikipedia vocabulary), as measured by the percentage of tokens in a testing set for which the language model could not predict the term embedding. As a testing set, we used the corpus reported in \citet{amram2018representations}.
  \end{enumerate}
  \item Downstream classification tasks: 
  \begin{enumerate}
    \item Named-entity recognition (NER) - the ability of the model to classify named entities in text, such as persons' names, organizations, and locations; tested on a labeled dataset from \citet{mordecai81hebrew}, and evaluated with F1-score ($2\times\frac{precision\times recall}{precision+recall}$).
    \item Part of speech (POS) - the ability of the model to classify the grammatical role that a word or phrase plays in a sentence (e.g., noun, pronoun, verb); tested on a labeled Israeli newspaper dataset from \citet{sima2001building}, and evaluated with F1-score (See appendix A for detailed explanation of how we compared the performance of the models).
    \item Polarity analysis - tested on the polarity data that were collected and labeled by \citet{amram2018representations} and evaluated with F1-score.
  \end{enumerate}
\end{enumerate}
The results of this comparison are shown in Table \ref{table:compare_language_model}. 
\begin{table}[h]
\centering
\begin{tabular}{lrrrrr}
\toprule
\textbf{}     & Fill-the-blank    & OOV       & NER         & POS         & Polarity analysis  \\
Metric      & \small \textit{(Perplexity)} & \small \textit{(\%)}  & \small \textit{(F-1 score)} & \small \textit{(F-1 score)} & \small \textit{(F-1 score)} \\
\midrule
chars  (1k)   & \textbf{1.17}     & \textbf{$\sim$0} & 0.74         & 0.92         & 0.69         \\
n-gram (30k) & 4.4          & \textbf{$\sim$0} & 0.79         & 0.90    & \textbf{0.79}    \\
n-gram  (50k) & 5.7          & \textbf{$\sim$0} & 0.79         & 0.92    & 0.71         \\
morpheme-based  & 8.9          & 0.75       & \textbf{0.92}         & \textbf{0.95}         & 0.65         \\
word-based    & 209830        & 50        & 0.86         & N/A$^*$          & 0.43        \\
\bottomrule
\end{tabular}
  \caption{Comparison of task performance for different input alternatives. The best results for each task are in bold.}
  \label{table:compare_language_model}
  \begin{tablenotes}
   \small
    \item $^*$ POS cannot be computed due to the high OOV percentage.
  \end{tablenotes}
\end{table}

For the unsupervised tasks (fill-in-the-blank and OOV), the sub-word representations (n-gram-based and morpheme-based) performed similarly well, and substantially outperformed the word-based representation. Specifically, all sub-word methods were able to capture OOV tokens, with the exception of special emojis. The performance for the fill-in-the-blank task is monotonic with respect to the dictionary size. Char-based representation outperformed the sub-word representations on the fill-in-the-blank task and performed equally well on the OOV task. This is not surprising, as smaller-sized dictionaries leave less room for mistakes.

For each of the downstream tasks, in line with our hypothesis, the top-performing tokenization was a sub-word representation. Specifically, n-gram-based tokenization performed best on the POS task, whereas morpheme-based tokenization achieved the best performance on NER. For polarity analysis, the n-gram-based approach with the smaller dictionary (30K) performed significantly better than all other approaches. 

These results suggest that (1) there is no single representation that is optimal for the entire set of tasks, (2) for each task, there is at least one sub-word representation that outperforms both the char- and word-based representations, and (3) for a sentiment analysis task, which is the focus of this work, an n-gram-based sub-word representation with a smaller dictionary yielded the highest performance. On the basis of these results, we selected the latter tokenization for our model. 

\subsection{Final Model}
In line with the specifications outlined above, we trained a large-size BERT on both the Wikipedia corpus and an OSCAR corpus \citep{ortiz-suarez-etal-2020-monolingual}, with a small-size n-gram-based sub-word dictionary. For the Hebrew language, OSCAR contains a corpus of size 9.8 GB, including 1 billion words and over 20.8 million sentences (after de-duplicating the original data). 
We used a Pytorch implementation of Transformers in Python \citep{wolf-etal-2020-transformers} to train a base-BERT network for 4 epochs, with learning rate = 5e-5, using the Adam optimizer in batches of 128 sentences each. 

The performance of the final model is reported in Table \ref{table:compare_heBERT_performances}, and compared to the performance of (i) the (non-BERT) model reported in \citet{amram2018representations} and \citet{more2019joint} and \citet{bareket2020neural}, the only other model developed for NLP tasks in Hebrew (denoted SOTA, or “state of the art”); and (ii) mBERT. The best results for each task are in bold. 
\begin{table}[h]
  \centering
  \begin{tabular}{lrrrrr}
    \toprule
    Task     & Fill-in-the-blank    & OOV       & NER         & POS         & Polarity analysis  \\
    Metric    & \small \textit{(Perplexity)} & \small \textit{(\%)}  & \small \textit{(F-1 score)} & \small \textit{(F-1 score)} & \small \textit{(F-1 score)} \\
    \midrule
    HeBERT    & 3.24     & \textbf{$\sim$0} & \textbf{0.85}    & 0.92    & \textbf{0.93}    \\
    Current SOTA & \textit{(Not reported)}    & 8\%   & 0.84         & \textbf{0.97}      & 0.89         \\
    mBERT    & \textbf{1}           & \textbf{0}    & 0.74         & 0.94    & 0.88         \\
    \bottomrule
  \end{tabular}
  \caption{HeBERT performance, compared to alternative models. Best results for each task are in bold. }
  \label{table:compare_heBERT_performances}
\end{table}

The results show that while mBERT outperformed HeBERT in an unsupervised task (fill-in-the-blank), HeBERT performed better on supervised tasks, even when compared to the current SOTA. Of note, mBERT contains only 2,000 tokens in Hebrew (compared to 30K in HeBERT). HeBERT’s higher performance in supervised tasks is thus not surprising. 

\section{HebEMO: A Model for Polarity Analysis and Emotion Recognition}
In this section we develop HebEMO - a model for sentiment analysis, including polarity analysis and emotion recognition. HebEMO, which is based on HeBERT, predicts sentiments at a document level; as elaborated in what follows, in our case a “document” is a single user-generated comment on a news website. The development of the model is based on three main elements: (i) data collection; (ii) data annotation; and (iii) fine-tuning of HeBERT. 

\subsection{Data Collection and Annotation}
The data collected for this study were compiled from user comments that were posted to Israeli news websites in response to Covid-19-related articles, during the Covid-19 pandemic (Jan-Dec, 2020) - a highly emotional period \citep{pedrosa2020emotional}. 

Our selection of news sites was inspired by a 2016 statement by Israel’s president, Reuven (Rubi) Rivlin, according to which Israeli society is composed of four equal-sized “tribes” which are culturally different (and hence might express emotions slightly differently); of these, three comprise Hebrew-speaking Jews - namely, secular, national-religious, and ultra-Orthodox (“Haredi”) - and the fourth “tribe” is Israel’s Arab population \citep{steiner2016president}. Each group is represented in both politics and the media. 

Accordingly, we collected data from three popular Israeli news sites that, respectively, represent the three Hebrew-speaking “tribes”. Specifically, our dataset contained all Covid-19-related articles from \textit{Ynet\footnote{https://www.ynet.co.il/}}, which is identified with the secular “tribe” (with a slight left-wing political leaning); \textit{Israel Hayom\footnote{https://www.israelhayom.co.il/}} (translation: “Israel Today”), which is identified with the national-religious “tribe” (with a slight right-wing political leaning), and \textit{Be-Hadre Haredim\footnote{https://www.bhol.co.il/}} (translation: “In Haredis' Rooms”), which represents the ultra-Orthodox group. 

For each article, we collected the article’s text, its date of publication, the section in the news site in which it was published (e.g., news, health, sports), the author, and the comments section. We excluded from the dataset comments that did not contain Hebrew words, and comments with fewer than 3 words. We further merged repeated consecutive characters (e.g., three or more identical punctuation symbols) and removed links and double spaces. The compiled corpus, summarized in Table \ref{table:Amount_of_talkbacks_for_website}, contained over half a million comments on 10,794 titles in various sections.
\begin{table}[h]
  \centering
  \resizebox{!}{1.5in}{%
  \begin{tabular}{llrr}
    \toprule
    source & section & \#titles & \#comments \\
    \midrule
    Ynet & activism &   3 &     80 \\
       & article &  500 &   30053 \\
       & articles &  4506 &   156174 \\
       & dating &   34 &    793 \\
       & digital &  124 &    2067 \\
       & economy &  1486 &   55521 \\
       & entertainment &  153 &    7000 \\
       & food &   41 &    4064 \\
       & health &  271 &    8112 \\
       & judaism &   63 &    4321 \\
       & news &  138 &    8214 \\
       & sport &   71 &    363 \\
       & vacation &  181 &   10467 \\
       & wheels &   32 &    964 \\
    Israel Hayom & article &  2651 &   71372 \\
       & opinion &   36 &    392 \\
    Be-Hadre Haredim & news &  191 &    805 \\
    \bottomrule
  \end{tabular}%
  }
  \caption{Description of the collected data}
   \label{table:Amount_of_talkbacks_for_website}
  \centering
\end{table}

\subsection{Data Annotation}
\begin{figure}
  \centering
  \resizebox{\columnwidth}{!}{  \includegraphics{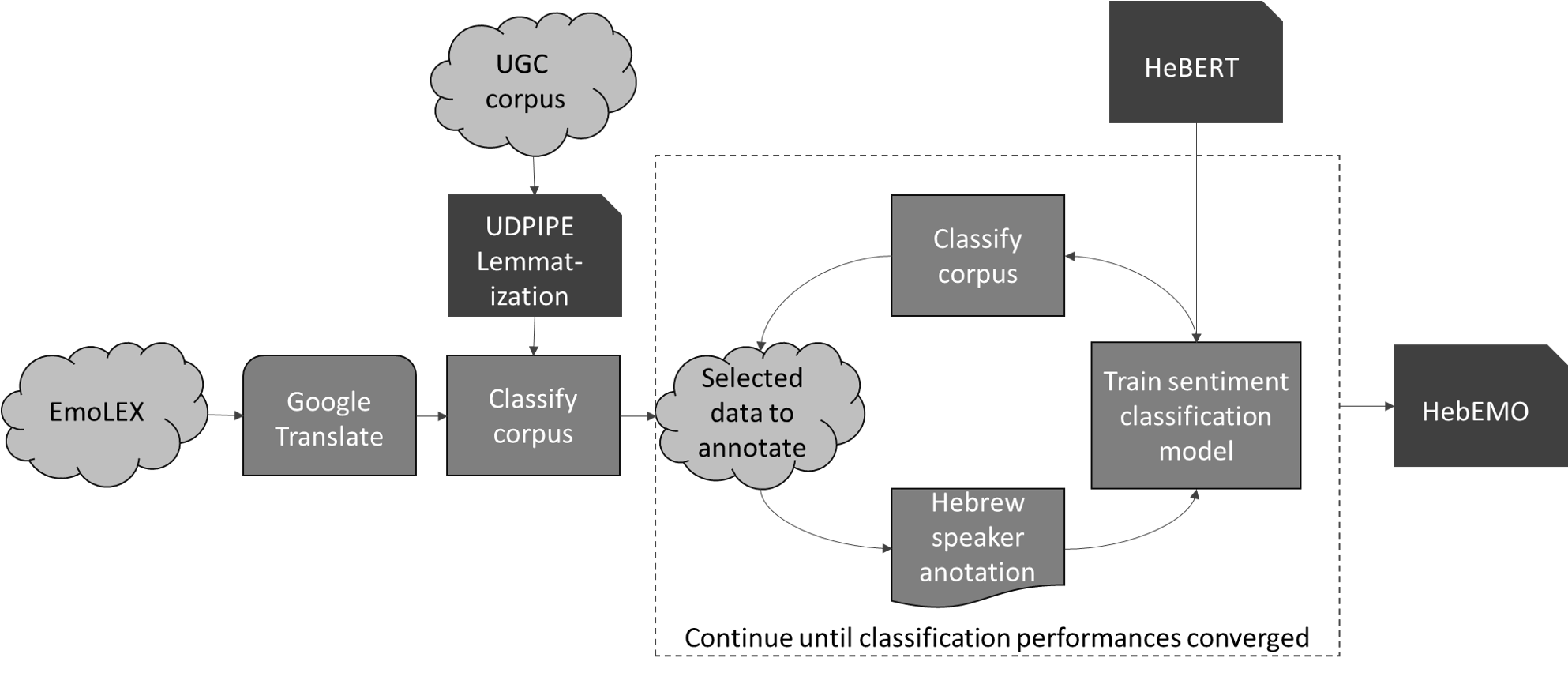}
  }
  \caption{Iterative annotation process}
  \label{fig:Annotation_and_Classification_Circle}
\end{figure}

We annotated a total of 4,000 comments. Comments were selected for annotation following active learning principles \citep{li2012active} to minimize the well-known imbalance problem in the emotion recognition literature \citep{acheampong2020text}. The annotation process we used is described below and illustrated in Figure \ref{fig:Annotation_and_Classification_Circle}.

Our iterative process was initialized in \textbf{step 1} with a naive unsupervised lexicon-based approach. For this step, we Google-translated EmoLex: a freely-available English-language polarity and emotion dictionary \citep{Mohammad13}. EmoLex contains a list of manually collected (via crowdsourcing) English words classified according to one or more of the eight basic emotions and two polarity values (positive and negative). We then used the translated dictionaries to score the entire set of \textit{lemmatized} comments in our dataset. Lemmatization was achieved with UDPipe \citep{straka2016udpipe}. 

In \textbf{step 2}, given the initial sentiment scores generated in step 1, we selected a set of 150 comments, of which 75 comments had received the highest positive polarity scores, and 75 had received the highest negative polarity scores. Similarly, for each of the eight emotions, we selected a set of 75 comments in which the emotion was highly expressed, and another 75 comments in which the emotion was not expressed. The resulting set, after removing duplicate comments, comprised a total of 1,500 initially labeled comments.

We then turned to Prolific\footnote{https://www.prolific.co/}, a trusted online labor and research platform, to manually re-annotate the 1,500 comments. Each comment was annotated by at least three distinct native Hebrew-speaking Prolific workers. Specifically, annotators were asked to rate individual comments’ polarity on a symmetric 5-point scale of \{strongly negative, negative, neutral, positive, strongly positive\}, and to rate the expression of each emotion in the comment on a polar 3-point scale of \{not expressed (in the comment), expressed, strongly expressed\}. The participants were given the context of the comment (i.e., the title of the news article on which the comment was posted). Each participant annotated 20 randomly selected comments.

The reliability levels of workers’ annotations were then computed with Krippendorff's alpha \citep{krippendorff1970estimating}, a measure of inter-rater agreement. We measured reliability independently for each sentiment in a comment, using coarser sentiment scales of polarity = \{positive, neutral, negative\} and emotion = \{expressed, not expressed\}. For example, if two raters, $i$ and $j$, rated the emotion “anger” in a comment $c$ as $L_{c, anger}^i$ = “expressed” and $L_{c, anger}^j$ = “strongly expressed”, we computed their mutual response as “agreement” (formally, the observed agreement between the raters was $\delta(L_{c,anger}^{i},L_{c,anger}^{j}) = 0$). If the ratings were $L_{c, anger}^i$ = “expressed” (or “strongly expressed”) and $L_{c, anger}^j$ = “not expressed”, we computed the raters’ mutual response as “disagreement” ($\delta(L_{c,anger}^{i},L_{c,anger}^{j}) = 1$). We then excluded comments’ sentiment annotation with Krippendorff's alpha lower than 0.75.

In \textbf{step 3}, we trained an initial HeBERT-based sentiment (supervised) classifier (see details in Section \ref{classification}) on the crowd-annotated data, and predicted polarity and emotion scores for the remainder of the corpus. 
We then repeated steps 2 and 3 until the performance of our classifier converged. Convergence occurred after three iterations, and a total of 4,000 partially labeled comments (partially means that the raters agreed on at least one sentiment). Tables \ref{tab:polarity_ratio} and \ref{tab:label_ratio} summarize the number of comments for each sentiment (polarity and emotion, respectively) for which there was high agreement among raters, and the percentage of the comments that express this sentiment. For example, the expression/non-expression of the emotion “anger” was labelled in 1,979 distinct comments; among these, “anger” was expressed in 78\% of the comments, and in 22\% it was not expressed. 

\begin{table}[h]
  \centering
  \begin{tabular}{lrr}
    \toprule
    Polarity  & \# labeled comments & \% comments  \\
    \midrule
    Positive & 253  & 13.8\% \\
    Neutral & 55  & 3\% \\
    Negative & 1525 & 83.2\% \\
    \bottomrule
  \end{tabular}
  \caption{Summary of the polarity data}
  \label{tab:polarity_ratio}
\end{table}
\begin{table}[h]
  \centering
  \begin{tabular}{lrr}
  \toprule
    Emotion & \# labeled comments & \% comments \\
    \midrule
    Anger    & 1979        & 78\%         \\
    Disgust   & 2115        & 83\%         \\
    Anticipation & 681         & 58\%         \\
    Fear    & 1041        & 45\%         \\
    Joy    & 2342        & 12\%         \\
    Sadness   & 998         & 59\%         \\
    Surprise  & 698         & 17\%         \\
    Trust    & 1956        & 11\%         \\
    \bottomrule
  \end{tabular}
  \caption{Summary of the emotion data}
  \label{tab:label_ratio}
\end{table}

Interestingly, though we attempted to balance the expression and non-expression of each sentiment in our labelled data, our raters had significantly lower agreement on positive sentiments - specifically, positive polarity, expression of happiness, surprise, and trust, and non-expression of anger and disgust.
In line with the theory of \citet{plutchik1980general}, we observed high negative correlation between emotions that are located opposite each other in Plutchik’s wheel of emotion, and positive correlation between closely related emotions (see Table \ref{table:correlation_among_emotions}).
\begin{table}[h]
  \centering
  \resizebox{\columnwidth}{!}{%
  \begin{tabular}{lrrrrrrrrr}
    \toprule
      & Anger & Disgust & Anticipation & Fear & Joy & Sadness & Surprise & Trust & Polarity \\
      \midrule
      Anger    & 1.00 &     &       &   &    &     &     &    &      \\
      Disgust   & 0.46 & 1.00  &       &   &    &     &     &    &      \\
      Anticipation & 0.10 & 0.09  & 1.00    &   &    &     &     &    &      \\
      Fear    & 0.15 & 0.11  & 0.14    & 1.00 &    &     &     &    &      \\
      Joy    & 0.25 & 0.27  & 0.12    & 0.11 & 1.00 &     &     &    &      \\
      Sadness   & 0.21 & 0.16  & 0.13    & 0.28 & 0.12 & 1.00  &     &    &      \\
      Surprise  & 0.06 & 0.04  & 0.10    & 0.15 & 0.05 & 0.12  & 1.00   &    &      \\
      Trust    & 0.27 & 0.31  & 0.11    & 0.07 & 0.41 & 0.08  & 0.07   & 1.00 &      \\
      \midrule
      Polarity  & 0.47 & 0.44  & 0.11    & 0.09 & 0.36 & 0.14  & 0.05   & 0.40 & 1.00   \\
      \bottomrule
    \bottomrule
  \end{tabular}
  }
  \caption{Pearson score for correlation among emotions identified by human raters}
  \label{table:correlation_among_emotions}
  \centering
\end{table}
The final classification model was denoted \textit{HebEMO}. 

\subsection{Fine-Tuning of HeBERT: The Classification Model} \label{classification}
We modeled our classification algorithm by fine-tuning HeBERT for a document-level classification task. Prediction probabilities were computed with a softmax activation function. We treated the polarity task as a multinomial problem with three classes (positive, neutral, negative); emotions were modeled as independent dichotomous classification tasks (expressed, not expressed), as multiple emotions can co-exist in a single comment. Attempts to merge emotion pairs (e.g., joy-sad) into a single classification category yielded lower performance. 
To train and evaluate our model, we randomly partitioned the corpus into training (70\%), validation (15\%), and test (15\%) sets. In order to avoid data leakage, the tokenization process (in HeBERT) was not trained on the UGC dataset. We repeated the training and evaluation process following a bootstrap approach with 50 samples (each generated a different data partition) and examined the stability of our results.

\section{Results}
We applied HebEMO to our annotated dataset and examined its performance, as measured by precision, recall, F1-score and overall accuracy of the expressed sentiment. Table \ref{tab:classification_model_performances_polarity} presents the performance of our model on the polarity task, and Table \ref{tab:classification_model_performances} presents the performance for emotion recognition. The weighted average performance across all sentiments is F1-score = 0.931, and overall accuracy = 0.91. With the exception of the emotion \textit{“surprise”}, the performance of the model ranges between F1-score and accuracy of 0.78-0.97. These performance levels, as far as we know, exceed those of state-of-the-art English-language models for UGC emotion recognition \citep{ghanbari2019text, mohammad2018semeval}. 

The emotion \textit{“surprise”} is known to be hard to detect. As mentioned in \citet{zhou2020emotion}, the best reported F1-score for this emotion in English was found to be as low as 0.19 \citep{mohammad2018semeval}. In our dataset, the amount of labeled data for “surprise” - as well as for its opposing counterpart on the wheel of emotion, “anticipation” \citep{plutchik1980general} - was also the lowest among all emotions (see Table \ref{tab:label_ratio}), implying that this pair is a challenging labeling task even for human annotators. 
\begin{table}[]
  \centering
  \begin{tabular}{llll}
  \toprule
         & Precision & Recall & F1-score \\
  \midrule
  Positive   & 0.96   & 0.92  & 0.94   \\
  Neutral   & 0.83   & 0.56  & 0.67   \\
  Negative   & 0.97   & 0.99  & 0.98   \\
  Accuracy   &      &    & 0.97   \\
  \bottomrule
  \end{tabular}
  \caption{HebEMO performance on polarity task in the UGC data}
  \label{tab:classification_model_performances_polarity}
\end{table}
\begin{table}[h]
  \centering
  \begin{tabular}{lrrrr}
    \toprule
          & F1  & Precision & Recall & Accuracy \\
    \midrule
    Anger    & 0.97 & 0.97   & 0.97  & 0.95   \\
    Disgust   & 0.96 & 0.97   & 0.95  & 0.93   \\
    Anticipation & 0.85 & 0.83   & 0.87  & 0.84   \\
    Fear    & 0.80 & 0.84   & 0.77  & 0.80   \\
    Joy     & 0.88 & 0.89   & 0.87  & 0.97   \\
    Sadness   & 0.84 & 0.83   & 0.84  & 0.79   \\
    Surprise  & 0.41 & 0.47   & 0.37  & 0.78   \\
    Trust    & 0.78 & 0.88   & 0.70  & 0.95   \\
    \bottomrule
  \end{tabular}
  \caption{HebEMO performance on emotion detection task in the UGC data}
  \label{tab:classification_model_performances}
\end{table}

Next, we re-trained HebEMO on the polarity data reported by \citet{amram2018representations}. \citet{amram2018representations} collected comments that were written in response to official tweets posted by the Israeli president, Mr. Reuven Rivlin, between June and August, 2014 (a total of 12,804 Hebrew comments). The authors manually annotated the comments with the following labels - supportive (positive), criticizing (negative), or off-topic (neutral) comments - and published a partitioned dataset (training and validation) for the benefit of comparisons between language models. 

The performance of our model is presented in Table \ref{table:tsarfati_polarity_analysis}, along with the improvement/ deterioration in performance as compared with the SOTA model reported in  \citet{amram2018representations}. The results show that, in most aspects, with the exception of off-topic precision, our model’s performance exceeds that of the SOTA model. The improvement is significant at the 95\% confidence level. 
\par
\begin{table}[h]
  \centering
  \begin{tabular}{cccc}
  
  \toprule
           & Precision                       & Recall                        & F1-score                        \\
    \midrule
    Positive  & \begin{tabular}[c]{@{}l@{}}0.95\\ \tiny (+.03)\end{tabular} & \begin{tabular}[c]{@{}l@{}}0.96\\ \tiny(+.01)\end{tabular} & \begin{tabular}[c]{@{}l@{}}0.95\\ \tiny(+.01)\end{tabular} \\
    Negative   & \begin{tabular}[c]{@{}l@{}}0.89\\ \tiny(+.05)\end{tabular} & \begin{tabular}[c]{@{}l@{}}0.89\\ \tiny(+.02)\end{tabular} & \begin{tabular}[c]{@{}l@{}}0.89\\ \tiny(+.04)\end{tabular} \\
    Off-topic  & \begin{tabular}[c]{@{}l@{}}0.70\\ \tiny(-.3)\end{tabular} & \begin{tabular}[c]{@{}l@{}}0.56\\ \tiny(+.55)\end{tabular} & \begin{tabular}[c]{@{}l@{}}0.62\\ \tiny(+.03)\end{tabular}  \\
    Accuracy   &                            &                            & \begin{tabular}[c]{@{}l@{}}0.93\\ \tiny(+.03)\end{tabular} \\
    \bottomrule
  \end{tabular}
  \caption{The performance of HebEMO when trained on the polarity corpus reported by \citet{amram2018representations}}
\label{table:tsarfati_polarity_analysis}
\end{table}
\par

\section{Summary and Discussion}
This paper presented two new tools that contribute to the development of Hebrew-language sentiment analysis capabilities: (i) \textbf{HeBERT} - the first Hebrew BERT model, and a new state-of-the-art model for multiple Hebrew NLP tasks; and (ii) \textbf{HebEMO} - a tool for polarity analysis and emotion recognition from Hebrew UGC. 

Although HeBERT was developed for the purpose of optimizing sentiment analysis, we showed that it outperforms mBERT in a variety of supervised language tasks. This finding is consistent with the literature that proposes that language-specific models are better than multilingual model. HeBERT also showed better performance than the current (non-BERT) SOTA Hebrew-language model.

For the task of extracting sentiments from UGC, we showed that a morpheme-based model, which aims to “understand” features of the language, performed less well than a model that did not address the language features (ngram-based sub-words). For the latter input representation, a smaller-size dictionary was better than the larger-size dictionary. A plausible explanation for these results is that UGC contains unofficial language, which includes non-lexical words such as slang words and typos. Over-fitting a model to a language in this case may overlook the unique characteristics of the unofficial language.
In future work we plan to examine the performance of HebEMO when HeBERT is trained on a PMI masking task, rather than fill-in-the-blank.
\newpage
\bibliographystyle{informs2014}
\bibliography{references}

\begin{thebibliography}{74}
\providecommand{\natexlab}[1]{#1}
\providecommand{\url}[1]{\texttt{#1}}
\providecommand{\urlprefix}{URL }

\bibitem[{Acheampong et~al.(2020)Acheampong, Wenyu, \protect\BIBand{}
  Nunoo-Mensah}]{acheampong2020text}
Acheampong FA, Wenyu C, Nunoo-Mensah H (2020) Text-based emotion detection:
  Advances, challenges, and opportunities. \emph{Engineering Reports} e12189.

\bibitem[{Adamopoulos et~al.(2018)Adamopoulos, Ghose, \protect\BIBand{}
  Todri}]{adamopoulos2018impact}
Adamopoulos P, Ghose A, Todri V (2018) The impact of user personality traits on
  word of mouth: Text-mining social media platforms. \emph{Information Systems
  Research} 29(3):612--640.

\bibitem[{Ahmad et~al.(2020)Ahmad, Jindal, Ekbal, \protect\BIBand{}
  Bhattachharyya}]{ahmad2020borrow}
Ahmad Z, Jindal R, Ekbal A, Bhattachharyya P (2020) Borrow from rich cousin:
  transfer learning for emotion detection using cross lingual embedding.
  \emph{Expert Systems with Applications} 139:112851.

\bibitem[{Ahorsu et~al.(2020)Ahorsu, Lin, Imani, Saffari, Griffiths,
  \protect\BIBand{} Pakpour}]{ahorsu2020fear}
Ahorsu DK, Lin CY, Imani V, Saffari M, Griffiths MD, Pakpour AH (2020) The fear
  of covid-19 scale: development and initial validation. \emph{International
  journal of mental health and addiction} .

\bibitem[{Amram et~al.(2018)Amram, David, \protect\BIBand{}
  Tsarfaty}]{amram2018representations}
Amram A, David AB, Tsarfaty R (2018) Representations and architectures in
  neural sentiment analysis for morphologically rich languages: A case study
  from modern hebrew. \emph{Proceedings of the 27th International Conference on
  Computational Linguistics}, 2242--2252.

\bibitem[{Antoun et~al.(2020)Antoun, Baly, \protect\BIBand{}
  Hajj}]{antoun2020arabert}
Antoun W, Baly F, Hajj H (2020) Arabert: Transformer-based model for arabic
  language understanding. \emph{arXiv preprint arXiv:2003.00104} .

\bibitem[{Argaman(2010)}]{argaman2010linguistic}
Argaman O (2010) Linguistic markers and emotional intensity. \emph{Journal of
  psycholinguistic research} 39(2):89--99.

\bibitem[{Bareket \protect\BIBand{} Tsarfaty(2020)}]{bareket2020neural}
Bareket D, Tsarfaty R (2020) Neural modeling for named entities and morphology
  (nemo\^{} 2). \emph{arXiv preprint arXiv:2007.15620} .

\bibitem[{Belinkov et~al.(2017)Belinkov, Durrani, Dalvi, Sajjad,
  \protect\BIBand{} Glass}]{belinkov2017neural}
Belinkov Y, Durrani N, Dalvi F, Sajjad H, Glass J (2017) What do neural machine
  translation models learn about morphology? \emph{arXiv preprint
  arXiv:1704.03471} .

\bibitem[{Bellstam et~al.(2020)Bellstam, Bhagat, \protect\BIBand{}
  Cookson}]{bellstam2020text}
Bellstam G, Bhagat S, Cookson JA (2020) A text-based analysis of corporate
  innovation. \emph{Management Science} .

\bibitem[{Bojanowski et~al.(2015)Bojanowski, Joulin, \protect\BIBand{}
  Mikolov}]{bojanowski2015alternative}
Bojanowski P, Joulin A, Mikolov T (2015) Alternative structures for
  character-level rnns. \emph{arXiv preprint arXiv:1511.06303} .

\bibitem[{Chatterjee et~al.(2019)Chatterjee, Narahari, Joshi, \protect\BIBand{}
  Agrawal}]{chatterjee2019semeval}
Chatterjee A, Narahari KN, Joshi M, Agrawal P (2019) Semeval-2019 task 3:
  Emocontext contextual emotion detection in text. \emph{Proceedings of the
  13th International Workshop on Semantic Evaluation}, 39--48.

\bibitem[{Chitturi et~al.(2007)Chitturi, Raghunathan, \protect\BIBand{}
  Mahajan}]{chitturi2007form}
Chitturi R, Raghunathan R, Mahajan V (2007) Form versus function: How the
  intensities of specific emotions evoked in functional versus hedonic
  trade-offs mediate product preferences. \emph{Journal of marketing research}
  44(4):702--714.

\bibitem[{Desmet \protect\BIBand{} Hoste(2013)}]{desmet2013emotion}
Desmet B, Hoste V (2013) Emotion detection in suicide notes. \emph{Expert
  Systems with Applications} 40(16):6351--6358.

\bibitem[{Devlin et~al.(2018)Devlin, Chang, Lee, \protect\BIBand{}
  Toutanova}]{devlin2018bert}
Devlin J, Chang MW, Lee K, Toutanova K (2018) Bert: Pre-training of deep
  bidirectional transformers for language understanding. \emph{arXiv preprint
  arXiv:1810.04805} .

\bibitem[{Dong et~al.(2014)Dong, Wei, Tan, Tang, Zhou, \protect\BIBand{}
  Xu}]{dong2014adaptive}
Dong L, Wei F, Tan C, Tang D, Zhou M, Xu K (2014) Adaptive recursive neural
  network for target-dependent twitter sentiment classification.
  \emph{Proceedings of the 52nd annual meeting of the association for
  computational linguistics (volume 2: Short papers)}, 49--54.

\bibitem[{Ekman(1999)}]{ekman1999basic}
Ekman P (1999) Basic emotions. \emph{Handbook of cognition and emotion}
  98(45-60):16.

\bibitem[{Fattah \protect\BIBand{} Fierke(2009)}]{fattah2009clash}
Fattah K, Fierke KM (2009) A clash of emotions: The politics of humiliation and
  political violence in the middle east. \emph{European journal of
  international relations} 15(1):67--93.

\bibitem[{Fedus et~al.(2018)Fedus, Goodfellow, \protect\BIBand{}
  Dai}]{fedus2018maskgan}
Fedus W, Goodfellow I, Dai AM (2018) Maskgan: Better text generation via
  filling in the\_. \emph{arXiv preprint arXiv:1801.07736} .

\bibitem[{Ghanbari-Adivi \protect\BIBand{} Mosleh(2019)}]{ghanbari2019text}
Ghanbari-Adivi F, Mosleh M (2019) Text emotion detection in social networks
  using a novel ensemble classifier based on parzen tree estimator (tpe).
  \emph{Neural Computing and Applications} 31(12):8971--8983.

\bibitem[{Hemmatian \protect\BIBand{} Sohrabi(2019)}]{hemmatian2019survey}
Hemmatian F, Sohrabi MK (2019) A survey on classification techniques for
  opinion mining and sentiment analysis. \emph{Artificial Intelligence Review}
  1--51.

\bibitem[{Hochreiter \protect\BIBand{} Schmidhuber(1997)}]{hochreiter1997long}
Hochreiter S, Schmidhuber J (1997) Long short-term memory. \emph{Neural
  computation} 9(8):1735--1780.

\bibitem[{Jawahar et~al.(2019)Jawahar, Sagot, \protect\BIBand{}
  Seddah}]{jawahar2019does}
Jawahar G, Sagot B, Seddah D (2019) What does bert learn about the structure of
  language?

\bibitem[{Joulin et~al.(2016)Joulin, Grave, Bojanowski, Douze, J{\'e}gou,
  \protect\BIBand{} Mikolov}]{joulin2016fasttext}
Joulin A, Grave E, Bojanowski P, Douze M, J{\'e}gou H, Mikolov T (2016)
  Fasttext.zip: Compressing text classification models. \emph{arXiv preprint
  arXiv:1612.03651} .

\bibitem[{Kim(2014)}]{kim2014convolutional}
Kim Y (2014) Convolutional neural networks for sentence classification.
  \emph{arXiv preprint arXiv:1408.5882} .

\bibitem[{Kim-Prieto \protect\BIBand{} Diener(2009)}]{kim2009religion}
Kim-Prieto C, Diener E (2009) Religion as a source of variation in the
  experience of positive and negative emotions. \emph{The Journal of Positive
  Psychology} 4(6):447--460.

\bibitem[{Klein \protect\BIBand{} Tsarfaty(2020)}]{klein2020getting}
Klein S, Tsarfaty R (2020) Getting the\#\# life out of living: How adequate are
  word-pieces for modelling complex morphology? \emph{Proceedings of the 17th
  SIGMORPHON Workshop on Computational Research in Phonetics, Phonology, and
  Morphology}, 204--209.

\bibitem[{K{\"o}vecses(2003)}]{kovecses2003metaphor}
K{\"o}vecses Z (2003) \emph{Metaphor and emotion: Language, culture, and body
  in human feeling} (Cambridge University Press).

\bibitem[{Kratzwald et~al.(2018)Kratzwald, Ili{\'c}, Kraus, Feuerriegel,
  \protect\BIBand{} Prendinger}]{kratzwald2018deep}
Kratzwald B, Ili{\'c} S, Kraus M, Feuerriegel S, Prendinger H (2018) Deep
  learning for affective computing: Text-based emotion recognition in decision
  support. \emph{Decision Support Systems} 115:24--35.

\bibitem[{Krippendorff(1970)}]{krippendorff1970estimating}
Krippendorff K (1970) Estimating the reliability, systematic error and random
  error of interval data. \emph{Educational and Psychological Measurement}
  30(1):61--70.

\bibitem[{Levine et~al.(2020)Levine, Lenz, Lieber, Abend, Leyton-Brown,
  Tennenholtz, \protect\BIBand{} Shoham}]{levine2020pmi}
Levine Y, Lenz B, Lieber O, Abend O, Leyton-Brown K, Tennenholtz M, Shoham Y
  (2020) Pmi-masking: Principled masking of correlated spans. \emph{arXiv
  preprint arXiv:2010.01825} .

\bibitem[{Li et~al.(2012)Li, Ju, Zhou, \protect\BIBand{} Lin}]{li2012active}
Li S, Ju S, Zhou G, Lin X (2012) Active learning for imbalanced sentiment
  classification. \emph{Proceedings of the 2012 Joint conference on empirical
  methods in natural language processing and computational natural language
  learning}, 139--148.

\bibitem[{Liu(2012)}]{liu2012sentiment}
Liu B (2012) Sentiment analysis and opinion mining. \emph{Synthesis lectures on
  human language technologies} 5(1):1--167.

\bibitem[{Liu \protect\BIBand{} Zhang(2012)}]{liu2012survey}
Liu B, Zhang L (2012) A survey of opinion mining and sentiment analysis.
  \emph{Mining text data}, 415--463 (Springer).

\bibitem[{Liu et~al.(2019)Liu, Shi, Ji, \protect\BIBand{} Jia}]{liu2019survey}
Liu R, Shi Y, Ji C, Jia M (2019) A survey of sentiment analysis based on
  transfer learning. \emph{IEEE Access} 7:85401--85412.

\bibitem[{Medhat et~al.(2014)Medhat, Hassan, \protect\BIBand{}
  Korashy}]{medhat2014sentiment}
Medhat W, Hassan A, Korashy H (2014) Sentiment analysis algorithms and
  applications: A survey. \emph{Ain Shams engineering journal} 5(4):1093--1113.

\bibitem[{Mohammad et~al.(2018)Mohammad, Bravo-Marquez, Salameh,
  \protect\BIBand{} Kiritchenko}]{mohammad2018semeval}
Mohammad S, Bravo-Marquez F, Salameh M, Kiritchenko S (2018) Semeval-2018 task
  1: Affect in tweets. \emph{Proceedings of the 12th international workshop on
  semantic evaluation}, 1--17.

\bibitem[{Mohammad \protect\BIBand{} Turney(2013)}]{Mohammad13}
Mohammad SM, Turney PD (2013) Crowdsourcing a word-emotion association lexicon
  29(3):436--465.

\bibitem[{Mordecai \protect\BIBand{} Elhadad(2005)}]{mordecai81hebrew}
Mordecai NB, Elhadad M (2005) Hebrew named entity recognition. \emph{MONEY}
  81(83.93):82--49.

\bibitem[{More et~al.(2019)More, Seker, Basmova, \protect\BIBand{}
  Tsarfaty}]{more2019joint}
More A, Seker A, Basmova V, Tsarfaty R (2019) Joint transition-based models for
  morpho-syntactic parsing: Parsing strategies for mrls and a case study from
  modern hebrew. \emph{Transactions of the Association for Computational
  Linguistics} 7:33--48.

\bibitem[{Ortiz~Su{\'a}rez et~al.(2020)Ortiz~Su{\'a}rez, Romary,
  \protect\BIBand{} Sagot}]{ortiz-suarez-etal-2020-monolingual}
Ortiz~Su{\'a}rez PJ, Romary L, Sagot B (2020) A monolingual approach to
  contextualized word embeddings for mid-resource languages. \emph{Proceedings
  of the 58th Annual Meeting of the Association for Computational Linguistics},
  1703--1714 (Online: Association for Computational Linguistics),
  \urlprefix\url{https://www.aclweb.org/anthology/2020.acl-main.156}.

\bibitem[{Ortony et~al.(1987)Ortony, Clore, \protect\BIBand{}
  Foss}]{ortony1987referential}
Ortony A, Clore GL, Foss MA (1987) The referential structure of the affective
  lexicon. \emph{Cognitive science} 11(3):341--364.

\bibitem[{Pan \protect\BIBand{} Yang(2009)}]{pan2009survey}
Pan SJ, Yang Q (2009) A survey on transfer learning. \emph{IEEE Transactions on
  knowledge and data engineering} 22(10):1345--1359.

\bibitem[{Patwa et~al.(2020)Patwa, Aguilar, Kar, Pandey, PYKL, Gamb{\"a}ck,
  Chakraborty, Solorio, \protect\BIBand{} Das}]{patwa2020semeval}
Patwa P, Aguilar G, Kar S, Pandey S, PYKL S, Gamb{\"a}ck B, Chakraborty T,
  Solorio T, Das A (2020) Semeval-2020 task 9: Overview of sentiment analysis
  of code-mixed tweets. \emph{arXiv e-prints} arXiv--2008.

\bibitem[{Pedrosa et~al.(2020)Pedrosa, Bitencourt, Fr{\'o}es, Cazumb{\'a},
  Campos, de~Brito, \protect\BIBand{} e~Silva}]{pedrosa2020emotional}
Pedrosa AL, Bitencourt L, Fr{\'o}es ACF, Cazumb{\'a} MLB, Campos RGB, de~Brito
  SBCS, e~Silva ACS (2020) Emotional, behavioral, and psychological impact of
  the covid-19 pandemic. \emph{Frontiers in psychology} 11.

\bibitem[{Pennebaker et~al.(2001)Pennebaker, Francis, \protect\BIBand{}
  Booth}]{pennebaker2001linguistic}
Pennebaker JW, Francis ME, Booth RJ (2001) Linguistic inquiry and word count:
  Liwc 2001. \emph{Mahway: Lawrence Erlbaum Associates} 71(2001):2001.

\bibitem[{Peters et~al.(2018)Peters, Neumann, Iyyer, Gardner, Clark, Lee,
  \protect\BIBand{} Zettlemoyer}]{peters2018deep}
Peters ME, Neumann M, Iyyer M, Gardner M, Clark C, Lee K, Zettlemoyer L (2018)
  Deep contextualized word representations. \emph{arXiv preprint
  arXiv:1802.05365} .

\bibitem[{Pfefferbaum \protect\BIBand{} North(2020)}]{pfefferbaum2020mental}
Pfefferbaum B, North CS (2020) Mental health and the covid-19 pandemic.
  \emph{New England Journal of Medicine} .

\bibitem[{Plutchik(1980)}]{plutchik1980general}
Plutchik R (1980) A general psychoevolutionary theory of emotion.
  \emph{Theories of emotion}, 3--33 (Elsevier).

\bibitem[{Pota et~al.(2019)Pota, Marulli, Esposito, De~Pietro,
  \protect\BIBand{} Fujita}]{pota2019multilingual}
Pota M, Marulli F, Esposito M, De~Pietro G, Fujita H (2019) Multilingual pos
  tagging by a composite deep architecture based on character-level features
  and on-the-fly enriched word embeddings. \emph{Knowledge-Based Systems}
  164:309--323.

\bibitem[{Radford et~al.(2018)Radford, Narasimhan, Salimans, \protect\BIBand{}
  Sutskever}]{radford2018improving}
Radford A, Narasimhan K, Salimans T, Sutskever I (2018) Improving language
  understanding by generative pre-training.

\bibitem[{Radford et~al.(2019)Radford, Wu, Child, Luan, Amodei,
  \protect\BIBand{} Sutskever}]{radford2019language}
Radford A, Wu J, Child R, Luan D, Amodei D, Sutskever I (2019) Language models
  are unsupervised multitask learners. \emph{OpenAI blog} 1(8):9.

\bibitem[{Rosaldo et~al.(1984)Rosaldo, Shweder, \protect\BIBand{}
  LeVine}]{rosaldo1984culture}
Rosaldo MZ, Shweder RA, LeVine RA (1984) Culture theory: essays on mind, self,
  and emotion.

\bibitem[{Schuster \protect\BIBand{} Nakajima(2012)}]{schuster2012japanese}
Schuster M, Nakajima K (2012) Japanese and korean voice search. \emph{2012 IEEE
  International Conference on Acoustics, Speech and Signal Processing
  (ICASSP)}, 5149--5152 (IEEE).

\bibitem[{Shapira et~al.(2020)Shapira, Lazarus, Goldberg, Gilboa-Schechtman,
  Tuval-Mashiach, Juravski, \protect\BIBand{} Atzil-Slonim}]{shapira2020using}
Shapira N, Lazarus G, Goldberg Y, Gilboa-Schechtman E, Tuval-Mashiach R,
  Juravski D, Atzil-Slonim D (2020) Using computerized text analysis to examine
  associations between linguistic features and clients’ distress during
  psychotherapy. \emph{Journal of counseling psychology} .

\bibitem[{Sima’an et~al.(2001)Sima’an, Itai, Winter, Altman,
  \protect\BIBand{} Nativ}]{sima2001building}
Sima’an K, Itai A, Winter Y, Altman A, Nativ N (2001) Building a tree-bank of
  modern hebrew text. \emph{Traitement Automatique des Langues} 42(2):247--380.

\bibitem[{Steiner(2016)}]{steiner2016president}
Steiner T (2016) President rivlin's" four tribes" initiative: The foreign
  policy implications of a democratic \& inclusive process to address israel's
  socio-demographic transformation .

\bibitem[{Straka et~al.(2016)Straka, Hajic, \protect\BIBand{}
  Strakov{\'a}}]{straka2016udpipe}
Straka M, Hajic J, Strakov{\'a} J (2016) Udpipe: trainable pipeline for
  processing conll-u files performing tokenization, morphological analysis, pos
  tagging and parsing. \emph{Proceedings of the Tenth International Conference
  on Language Resources and Evaluation (LREC'16)}, 4290--4297.

\bibitem[{Tay et~al.(2020)Tay, Dehghani, Bahri, \protect\BIBand{}
  Metzler}]{tay2020efficient}
Tay Y, Dehghani M, Bahri D, Metzler D (2020) Efficient transformers: A survey.
  \emph{arXiv preprint arXiv:2009.06732} .

\bibitem[{Tsarfaty et~al.(2020)Tsarfaty, Bareket, Klein, \protect\BIBand{}
  Seker}]{tsarfaty2020spmrl}
Tsarfaty R, Bareket D, Klein S, Seker A (2020) From spmrl to nmrl: What did we
  learn (and unlearn) in a decade of parsing morphologically-rich languages
  (mrls)? \emph{arXiv preprint arXiv:2005.01330} .

\bibitem[{Tsarfaty et~al.(2010)Tsarfaty, Seddah, Goldberg, K{\"u}bler, Versley,
  Candito, Foster, Rehbein, \protect\BIBand{} Tounsi}]{tsarfaty2010statistical}
Tsarfaty R, Seddah D, Goldberg Y, K{\"u}bler S, Versley Y, Candito M, Foster J,
  Rehbein I, Tounsi L (2010) Statistical parsing of morphologically rich
  languages (spmrl) what, how and whither. \emph{Proceedings of the NAACL HLT
  2010 First Workshop on Statistical Parsing of Morphologically-Rich
  Languages}, 1--12.

\bibitem[{Ullah et~al.(2016)Ullah, Amblee, Kim, \protect\BIBand{}
  Lee}]{ullah2016valence}
Ullah R, Amblee N, Kim W, Lee H (2016) From valence to emotions: Exploring the
  distribution of emotions in online product reviews. \emph{Decision Support
  Systems} 81:41--53.

\bibitem[{Vania et~al.(2018)Vania, Grivas, \protect\BIBand{}
  Lopez}]{vania2018character}
Vania C, Grivas A, Lopez A (2018) What do character-level models learn about
  morphology? the case of dependency parsing. \emph{arXiv preprint
  arXiv:1808.09180} .

\bibitem[{Vaswani et~al.(2017)Vaswani, Shazeer, Parmar, Uszkoreit, Jones,
  Gomez, Kaiser, \protect\BIBand{} Polosukhin}]{vaswani2017attention}
Vaswani A, Shazeer N, Parmar N, Uszkoreit J, Jones L, Gomez AN, Kaiser {\L},
  Polosukhin I (2017) Attention is all you need. \emph{Advances in neural
  information processing systems}, 5998--6008.

\bibitem[{Wierzbicka(1994)}]{wierzbicka1994emotion}
Wierzbicka A (1994) Emotion, language, and cultural scripts. .

\bibitem[{Wolf et~al.(2020)Wolf, Debut, Sanh, Chaumond, Delangue, Moi, Cistac,
  Rault, Louf, Funtowicz, Davison, Shleifer, von Platen, Ma, Jernite, Plu, Xu,
  Scao, Gugger, Drame, Lhoest, \protect\BIBand{}
  Rush}]{wolf-etal-2020-transformers}
Wolf T, Debut L, Sanh V, Chaumond J, Delangue C, Moi A, Cistac P, Rault T, Louf
  R, Funtowicz M, Davison J, Shleifer S, von Platen P, Ma C, Jernite Y, Plu J,
  Xu C, Scao TL, Gugger S, Drame M, Lhoest Q, Rush AM (2020) Transformers:
  State-of-the-art natural language processing. \emph{Proceedings of the 2020
  Conference on Empirical Methods in Natural Language Processing: System
  Demonstrations}, 38--45 (Online: Association for Computational Linguistics),
  \urlprefix\url{https://www.aclweb.org/anthology/2020.emnlp-demos.6}.

\bibitem[{Wu et~al.(2016)Wu, Schuster, Chen, Le, Norouzi, Macherey, Krikun,
  Cao, Gao, Macherey et~al.}]{wu2016google}
Wu Y, Schuster M, Chen Z, Le QV, Norouzi M, Macherey W, Krikun M, Cao Y, Gao Q,
  Macherey K, et~al. (2016) Google's neural machine translation system:
  Bridging the gap between human and machine translation. \emph{arXiv preprint
  arXiv:1609.08144} .

\bibitem[{Yadav \protect\BIBand{} Vishwakarma(2020)}]{yadav2020sentiment}
Yadav A, Vishwakarma DK (2020) Sentiment analysis using deep learning
  architectures: a review. \emph{Artificial Intelligence Review}
  53(6):4335--4385.

\bibitem[{Yin et~al.(2017)Yin, Kann, Yu, \protect\BIBand{}
  Sch{\"u}tze}]{yin2017comparative}
Yin W, Kann K, Yu M, Sch{\"u}tze H (2017) Comparative study of cnn and rnn for
  natural language processing. \emph{arXiv preprint arXiv:1702.01923} .

\bibitem[{Yue et~al.(2019)Yue, Chen, Li, Zuo, \protect\BIBand{}
  Yin}]{yue2019survey}
Yue L, Chen W, Li X, Zuo W, Yin M (2019) A survey of sentiment analysis in
  social media. \emph{Knowledge and Information Systems} 1--47.

\bibitem[{Zampieri et~al.(2019)Zampieri, Malmasi, Nakov, Rosenthal, Farra,
  \protect\BIBand{} Kumar}]{zampieri2019semeval}
Zampieri M, Malmasi S, Nakov P, Rosenthal S, Farra N, Kumar R (2019)
  Semeval-2019 task 6: Identifying and categorizing offensive language in
  social media (offenseval). \emph{arXiv preprint arXiv:1903.08983} .

\bibitem[{Zhang et~al.(2018)Zhang, Wang, \protect\BIBand{} Liu}]{zhang2018deep}
Zhang L, Wang S, Liu B (2018) Deep learning for sentiment analysis: A survey.
  \emph{Wiley Interdisciplinary Reviews: Data Mining and Knowledge Discovery}
  8(4):e1253.

\bibitem[{Zhong et~al.(2019)Zhong, Wang, \protect\BIBand{}
  Miao}]{zhong2019knowledge}
Zhong P, Wang D, Miao C (2019) Knowledge-enriched transformer for emotion
  detection in textual conversations. \emph{arXiv preprint arXiv:1909.10681} .

\bibitem[{Zhou et~al.(2020)Zhou, Wu, Wang, Xie, Tu, \protect\BIBand{}
  Li}]{zhou2020emotion}
Zhou D, Wu S, Wang Q, Xie J, Tu Z, Li M (2020) Emotion classification by
  jointly learning to lexiconize and classify. \emph{Proceedings of the 28th
  International Conference on Computational Linguistics}, 3235--3245.

\end{thebibliography}
\end{document}